\documentclass[cameraready]{Interspeech}
\title{ALARM: Audio–Language Alignment for Reasoning Models}

\author[affiliation={1, 2}, orcid=0009-0008-4480-5595]{Petr}{Grinberg}
\author[affiliation={2}]{Hassan}{Shahmohammadi}
\address{
    $^1$ EPFL, Switzerland \\
    $^2$ Sony Europe Ltd., Germany
}

\email{petr.grinberg@epfl.ch, Hassan.Shahmohammadi@sony.com}


\usepackage{comment}
\usepackage{tabularx}
\usepackage{multirow}
\usepackage{amssymb}

\usepackage{xcolor}

\begin{document}

\maketitle

\begin{abstract}
Large audio language models (ALMs) extend LLMs with auditory understanding. A common approach freezes the LLM and trains only an adapter on self-generated targets. However, this fails for reasoning LLMs (RLMs) whose built-in chain-of-thought traces expose the textual surrogate input, yielding unnatural responses. We propose self-rephrasing, converting self-generated responses into audio-understanding variants compatible with RLMs while preserving distributional alignment. We further fuse and compress multiple audio encoders for stronger representations. For training, we construct a 6M-instance multi-task corpus (2.5M unique prompts) spanning 19K hours of speech, music, and sound. Our 4B-parameter ALM outperforms similarly sized models and surpasses most larger ALMs on related audio-reasoning benchmarks, while preserving textual capabilities with a low training cost. Notably, we achieve the best open-source result on the MMAU-speech and MMSU benchmarks and rank third among all the models.
\end{abstract}

\section{Introduction}\label{sec:introduction}
The demonstrated success of large language models (LLMs) in text-based tasks~\cite{naveed2025comprehensive} has motivated the integration of additional modalities, such as vision~\cite{liu2023visual} and audio~\cite{fang2024llama}, as a natural step toward the development of general artificial intelligence. In this paper, we study the integration of audio input into state-of-the-art text-based LLMs. Such models, endowed with auditory understanding capabilities, are commonly referred to as large audio language models (LALMs), or audio language models (ALMs). Due to the scarcity of supervised audio–text data and the high computational cost of training, ALMs are typically built by equipping the text LLMs with an audio encoder for audio feature extraction and an adapter~\cite{lu2024desta, lu2025desta2, abouelenin2025phi, xu2025qwen2, tang2024salmonn}. 
However, fine-tuning LLMs with additional audio data is not only computationally expensive—even when using parameter-efficient fine-tuning (PEFT) techniques~\cite{han2024parameter}—but also often leads to degradation of the model’s text capabilities~\cite{hsiao2025analyzing, lu2025speech}, a phenomenon commonly referred to as catastrophic forgetting~\cite{kirkpatrick2017overcoming}. A practical remedy is to keep the LLM frozen during training~\cite{lu2025developing, fan2025alignformer} and to train only a modality adapter to bridge the modality gap. However, when target responses are human-annotated or generated by other LLMs, there is a mismatch between the distribution of target responses and that of the frozen pretrained LLM ~\cite{lu2025desta2}. Semantic and structural discrepancies between the target texts in the training data and the outputs of the frozen pretrained LLM introduce stylistic and lexical inconsistencies. As a result, ALM performance degrades, since the frozen LLM is forced to generate responses that lie outside its prior distribution, leading to reduced performance when performing audio-related tasks~\cite{lu2025desta2}.

The self-generation of target responses~\cite{lu2025desta2, lu2025developing} tackle this problem by implicitly aligning audio and text inputs. In this approach, a frozen LLM generates target responses $R$ conditioned on (1) textual representations $A_{\text{text}}$ of the corresponding audio inputs, such as speech transcriptions~\cite{fan2025alignformer} or metadata-enriched captions~\cite{lu2025desta2, lu2025developing}, and (2) instruction prompts $P$ that query the LLM. During training, $A_{\text{text}}$ is replaced with the corresponding audio signals $A_{\text{audio}}$, and a modality adapter is trained on input--target pairs of the form $\{A_{\text{audio}}, P\} \rightarrow R$. Since $R$ is generated by the model itself, this approach avoids the output distribution shift.

Nevertheless, the standard self-generation technique does not generalize to state-of-the-art LLMs with built-in chain-of-thought reasoning capabilities (RLMs)~\cite{guo2025deepseek, yang2025qwen3}. Specifically, replacing $A_{\text{audio}}$ with the corresponding $A_{\text{text}}$ is insufficient for RLMs, because their built-in reasoning process reveals the textual nature of the input, resulting in unnatural responses (see Table~\ref{tab:rephrasing_example}) during inference. To tackle this issue, we propose a self-rephrasing approach that converts generated responses $R$ into textual audio-understanding variants, while still avoiding output-distribution mismatch.

Moreover, many state-of-the-art self-generation-based ALMs~\cite{lu2024desta, lu2025desta2, lu2025developing} apply voice activity detection (VAD) and automatic speech recognition (ASR) to the audio, often together with additional audio features. However, explicitly relying on ASR can be problematic for general audio understanding due to its sensitivity to irrelevant speech-like signals and VAD errors. For example, background audio such as a TV broadcast or distant conversations in a park may spuriously activate the ASR module, even when the query concerns other aspects of the input. Similarly, VAD may not detect speech in low signal-to-noise (SNR) ratio scenarios leading to content loss. We eliminate the need for ASR-derived inputs and further strengthen the representation by integrating multiple audio encoders mostly operating at a 50~Hz token rate, while effectively compressing the resulting information from all encoders from 175~Hz to a length comparable to that of a single encoder at 25~Hz or 50~Hz. We achieve this by proposing three fusion methods based on cross-attention and Perceiver \cite{jaegle2022perceiver}.

Our work extends state-of-the-art approaches~\cite{lu2025desta2, lu2025developing} while addressing their critical limitations: (1) the incompatibility of standard self-generation with RLMs, (2) noise introduced by ASR-based inputs, and (3) hallucinations caused by artifacts in the constructed training data.

Our main contributions are as follows: 
\begin{itemize}
    \item We collect a multi-task corpus comprising 6M instances (2.5M unique prompts) spanning 19K hours of speech, music, and general audio. Our prompt generation and filtering pipeline promotes prompt diversity and audio--prompt alignment, reducing the risk of hallucinations observed in the prior large self-generated public release, DeSTA-AQA5M~\cite{lu2025desta2}, which contains only 7K hours of audio and 7K unique prompts.
    \item We extend the self-generated multimodal LLM paradigm to reasoning models and develop a multi-encoder ALM. Our 4B ALM, ALARM-E, outperforms models of similar size and even surpasses most larger ALMs on MMSU~\cite{wang2025mmsu}, ranking third among the other models, while preserving the backbone’s textual capabilities and requiring substantially lower training cost and amount of data. Morevoer, we achieve the best open-source result on MMAU~\cite{sakshi2025mmau}-speech benchmark and the top-3 with closed-source systems included.
    
    \item We remove the need to run ASR on the audio input and instead enhance the input representation via multiple encoders, enabling robust understanding of both vocal and non-vocal signals while providing a compressed token sequence.
    \item To facilitate further research, we open-source \footnote{\url{https://github.com/Blinorot/ALARM}} the code, including the data collection and generation scripts, along with our models' checkpoints.
\end{itemize}
\section{Methodology}\label{sec:methodology}
This section details our proposed approach, including the dataset construction pipeline and the technical design of our model.
\subsection{Dataset Construction}

\begin{figure*}[!tp]
    \centering
    \includegraphics[width=1.0\linewidth]{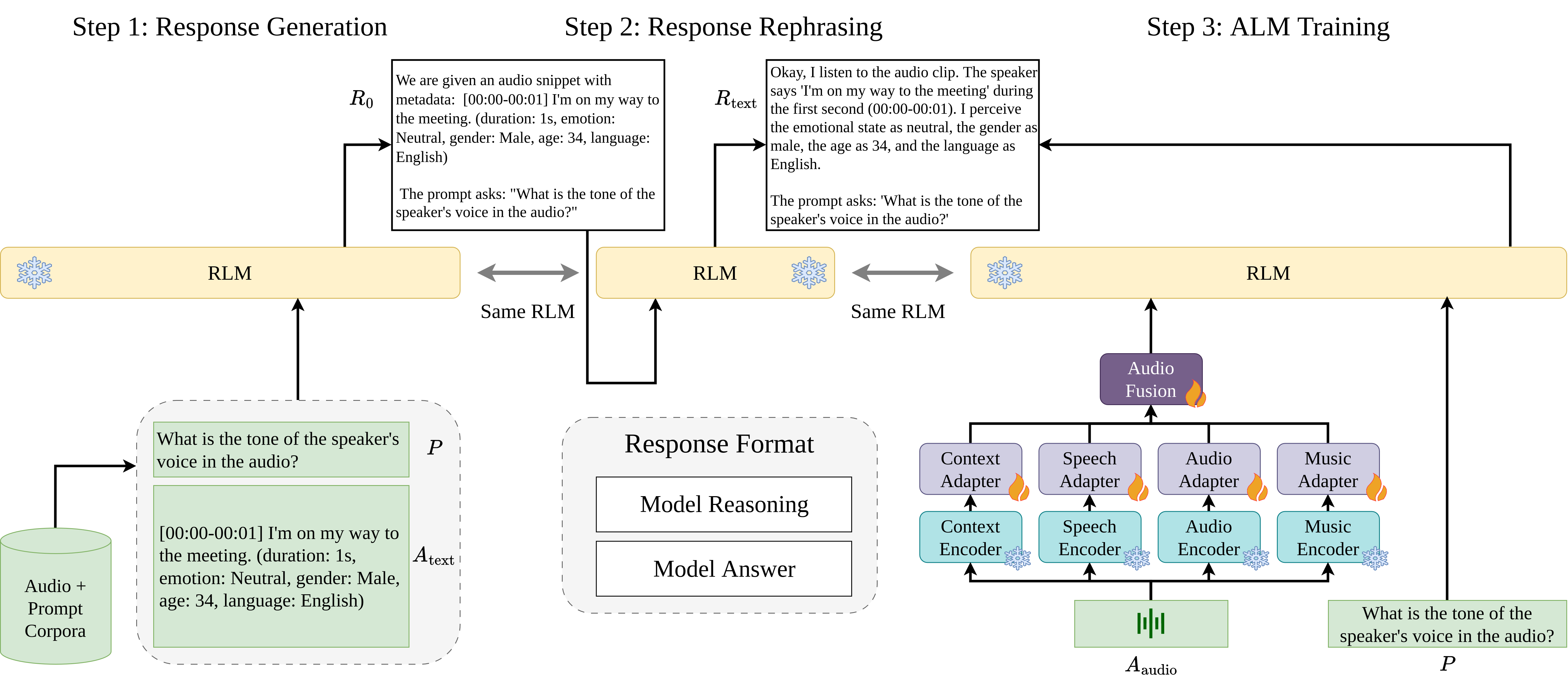}
    \caption{Overview of ALARM training pipeline. A pre-trained RLM first generates an initial response $R_0$ from textual metadata, then rephrases it into $R_{\text{text}}$. Finally, the RLM is equipped with the audio-fusion module and trained to generate $R_{\text{text}}$ from the corresponding audio inputs.}
    \label{fig:training_overview}
\end{figure*}

Let $\mathcal{D}_0=\{(A_{\text{audio}}^{i}, A_{\text{text}}^{i})\}_{i=1}^{N}$ denote a preliminary dataset, where $A_{\text{audio}}^{i}$ is an audio clip and $A_{\text{text}}^{i}$ is its associated textual description with metadata. Metadata fields can represent emotion, gender, nationality, audio quality, noisiness, or any other available information about audio (see Figure \ref{fig:training_overview} for an example).
Given $\mathcal{D}_0$, our goal is to construct the following dataset
\[
\mathcal{D}=\{(A_{\text{audio}}^{i}, P^{i}, R_{\text{text}}^{i})\}_{i=1}^{N},
\]
where $P^{i}$ is a prompt (instruction/question) about $A_{\text{audio}}^{i}$, and $R_{\text{text}}^{i}$ is the corresponding textual response from an LLM with reasoning capabilities (RLM).

Constructing $\mathcal{D}$ poses two main challenges. 
(1) Generating a diverse set of prompts $P$ while ensuring consistency with the corresponding metadata $A_\text{text}$. That is, each prompt $P^{i}$ should be answerable from $A_{\text{text}}^{i}$ (and thus from $A_{\text{audio}}^{i}$) without requiring information that is absent from the metadata. For example, asking about the noise level of the recording when $A_{\text{text}}^{i}$ does not mention the noise level leads to hallucinated responses. (2) The responses $R_{\text{text}}^{i}$ should not reveal the textual nature of the input. In practice, we observe that the RLM often repeats parts of the metadata in its reasoning trace (see Table 
\ref{tab:rephrasing_example}); when such outputs are used for training, the resulting model learns to produce responses that sound like they were derived from text rather than from audio, deviating from the natural behavior expected of audio-understanding models.

To address Challenge~(1), we use a fixed, pretrained instruction LLM, denoted by $Q$ (Qwen3-30B-A3B-Instruct-2507-FP8~\cite{yang2025qwen3}), as a prompt generator. For each sample $A_{\text{text}}^{i}$, we first sample a candidate set of prompts
\[
\tilde{\mathcal{P}^{i}}=\{P_{j}^{i}\}_{j=1}^{20},
\qquad P_{j}^{i} \sim \mathcal{Q}\left(\,\cdot \mid A_{\text{text}}^{i}, I\right),
\]
where $I$ denotes the instruction rules (i.e., the \textit{system prompt}) provided to $Q$. We design $I$ to enforce alignment between prompts and descriptions, while promoting multilingual coverage and prompt diversity\footnote{All instructions are publicly available in the shared repository}. We then apply $Q$ again to filter $\tilde{\mathcal{P}^{i}}$ to retain only prompts that (1) are answerable given the information in $A_{\text{text}}^{i}$ and (2) do not explicitly expose the textual nature of the input (e.g., phrases such as ``provided metadata'' or ``given description''). Finally, we select a \emph{single} training prompt by uniform sampling from the filtered set $\widetilde{\mathcal{P}}^{i}_{\text{filtered}}$:
\[
P^{i} \sim \mathrm{Unif}\left(\widetilde{\mathcal{P}}^{i}_{\text{filtered}}\right).
\]

\begin{figure}[!tp]
    \centering
    \includegraphics[width=1.0\linewidth]{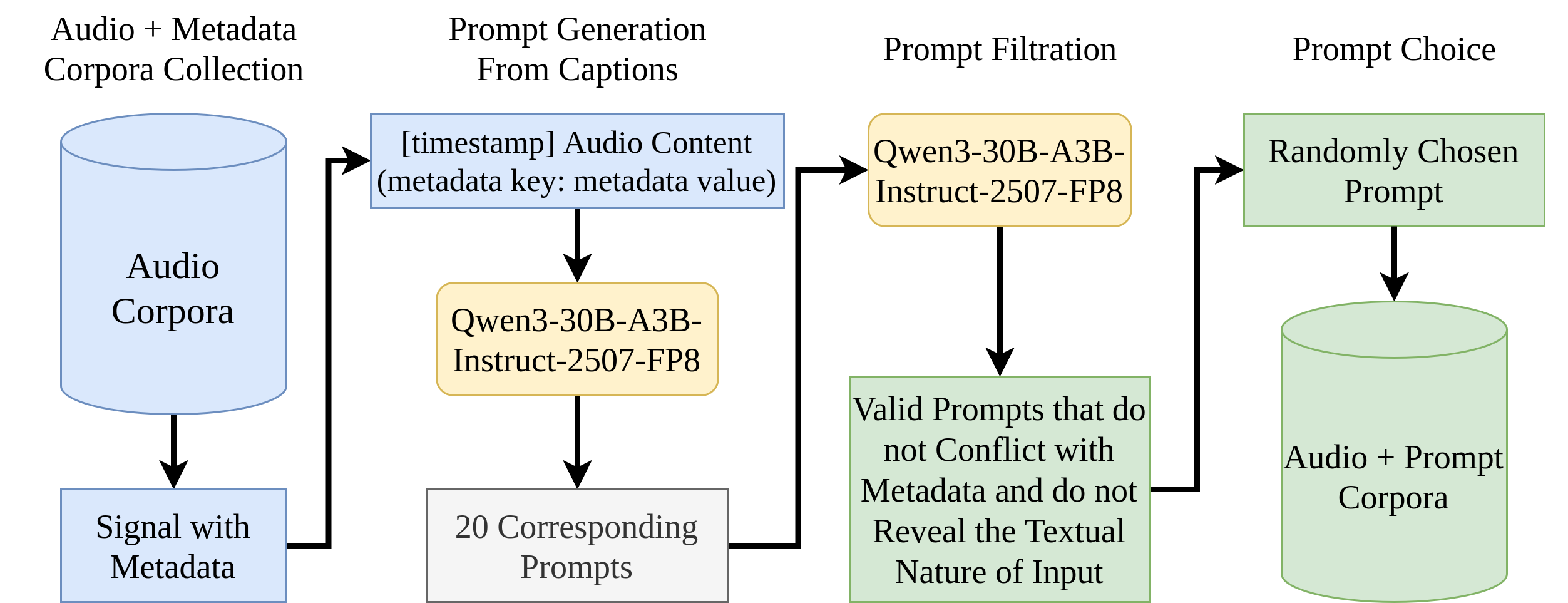}
    \caption{Overview of the corpora collection pipeline.}
    \label{fig:dataset_overview}
\end{figure}

Figure~\ref{fig:dataset_overview} illustrates the construction pipeline of the dataset $\mathcal{D}$. Compared to the data construction procedure in DeSTA-AQA5M~\cite{lu2025desta2}, their pipeline synthetically fills in missing metadata (e.g., speech transcriptions for VoxCeleb~\cite{nagrani2017voxceleb}, which are not included in the official release). This practice reduces reproducibility and can propagate artifacts---including annotator-induced hallucinations---into the resulting ALM. Moreover, their prompts $\mathcal{P}$ are sampled at random from a predefined pool of 7K options, without validating whether the available metadata $A_{\text{text}}$ contains sufficient information to answer the selected $P^{i}$.

\begin{table}[!tp]
    \centering
    \caption{Overview of our compiled dataset (total / train / validation) in comparison with DeSTA.}
    \label{tab:dataset_overview}
    \resizebox{\columnwidth}{!}{
    \begin{tabular}{c|ccc}
    \toprule
    Audio Type &  $\#$ Elements (M) & $\#$ Hours (K) & $\#$ Unique Prompts (M) \\
    \midrule
    Speech & $2.91/2.60/0.29$ & $9.88/8.83/0.98$ & $1.40/1.27/0.16$\\ 
    Sound & $2.01/1.80/0.20$ & $5.54/4.98/0.55$ & $0.36/0.33/0.06$\\ 
    Music & $0.59/0.53/0.06$ & $2.45/2.21/0.24$ & $0.16/0.14/0.03$\\
    \midrule
    Instruction & $0.57/0.56/0.01$ & $1.02/1.01/0.01$ & $0.57/0.56/0.01$\\ 
    \midrule
    Total & $6.08/5.49/0.56$ & $18.89/17.03/1.78$ & $2.49/2.3/0.26$\\ 
    \midrule
    DeSTA~\cite{lu2025desta2} & $4.96/4.96/0.00$ & $7.00/7.00/0.00$ & $0.007/0.007/0$\\
    \bottomrule
    \end{tabular}
    }
\end{table}

Next, we complete the construction of $\mathcal{D}$ by generating $R_{\text{text}}^{i}$ conditioned on $(A_{\text{text}}^{i}, P^{i})$. This step faces Challenge~(2): for RLMs, the reasoning trace often exposes that the model is operating on a textual description rather than raw audio (See Table~\ref{tab:rephrasing_example}, first row). Training on such outputs leads to unnatural responses and encourages the model to treat the audio modality as text. Adjusting the system prompt is ineffective, since the model often refers the rules in its reasoning. To preserve the pretrained capabilities of the RLM and avoid distribution shift at the same time, we keep the RLM model $Q_r$ frozen and generate natural targets via a two-stage self-rephrasing procedure. In stage 1, given $(A_{\text{text}}^{i}, P^{i})$, we sample an initial response
\[
R_{\text{0}}^{i} \sim \mathcal{Q}_r(\,\cdot \mid A_{\text{text}}^{i}, P^{i}).
\]

In stage 2 we prompt the \emph{same frozen} $Q_r$ to rewrite $R_{\text{0}}^{i}$ into an audio-grounded style (replacing text-based phrasing with auditory-perception ones), producing the final target
\[
R_{\text{text}}^{i} \sim \mathcal{Q}_r(\,\cdot \mid R_{\text{0}}^{i}, I_{\text{reph}}),
\]
where $I_{\text{reph}}$ specifies the rephrasing rules. Rephrasing is applied to both the reasoning and the final-answer parts of $R_{\text{0}}^{i}$. Because $R_{\text{text}}^{i}$ is generated by the same frozen $Q_r$, we avoid weight updates and mitigate output-distribution mismatch; moreover, the $Q_r$'s thinking process when rephrasing is kept separate from the returned $R_{\text{text}}^{i}$, allowing flexible control via $I_{\text{reph}}$ without leaking textual-input cues and system rules into the target response. As shown in the second row of Table~\ref{tab:rephrasing_example}, the rephrased response is more natural and treats audio as a distinct modality, which can improve audio understanding. However, RLM's reasoning traces are often lengthy, which is computationally expensive; we therefore enforce a thinking budget of $B=1536$ tokens during rephrasing; varying $B$ provides a direct quality--speed trade-off. As illustrated in the last row of Table~\ref{tab:rephrasing_example}, truncating the rephrasing chain-of-thought yields a slightly less rewritten $R_{\text{text}}$, while remaining perceptually aligned with the audio.

In our experiments, we use Qwen3-4B-Thinking-2507~\cite{yang2025qwen3} as the RLM ($Q_r$) to generate the final responses $R_{\text{text}}$. This model also serves as the backbone of our ALM. We make use of the \emph{training partitions} of the following datasets in our ALARM corpus, which consists of 6M total signals (19K hours) and 2.5M \emph{unique} prompts:
\begin{itemize}
    \item \textbf{Speech}. Cameo~\cite{christop2025cameo}, GlobeV3~\cite{wang2024globe}, VCTK~\cite{yamagishi2017vctk}, VocalSound~\cite{gong2022vocalsound}, ASCEND~\cite{lovenia2022ascend}, DisfluencySpeech~\cite{wang2024disfluencyspeech}, NISQA~\cite{mittag2021nisqa}, MUCS~\cite{diwan2021multilingual}, MLS~\cite{pratap2020mls}, LibriSpeech~\cite{panayotov2015librispeech}, ASVspoof19~\cite{wang2020asvspoof}, ASVspoof5~\cite{wang2024asvspoof}, Noisy VCTK~\cite{botinhao2016investigating, botinhao2016speech}.
    
    \item \textbf{Multi-speaker}. VoxCeleb 1~\cite{nagrani2017voxceleb}, Alimeeting~\cite{yu2022m2met}, AMI Corpus~\cite{carletta2005ami}, Meld~\cite{poria2019meld}, DailyTalk~\cite{lee2023dailytalk}, LibriMix~\cite{cosentino2020librimix}.
    
    \item \textbf{Music}. GTZAN~\cite{tzanetakis2002musical}, OpenSinger~\cite{huang2021multi}, SingMOS~\cite{tang2024singmos}, Mridangam~\cite{anantapadmanabhan2013modal}, Nsynth~\cite{engel2017neural}, FMA~\cite{defferrard2016fma}, SonicMaster~\cite{melechovsky2025sonicmaster}.
    
    \item \textbf{Sound}. ESC50~\cite{piczak2015esc}, FSD50K~\cite{fonseca2021fsd50k}, AudioSet~\cite{gemmeke2017audio}, AudioCaps~\cite{kim2019audiocaps}, Clotho~\cite{drossos2020clotho}.
\end{itemize}
We further supplement it with additional speech instructions data. HeySQuAD (human subset)~\cite{wu2023heysquad} is repeated twice (145K examples total): (1) questions and (2) questions with additional textual context based on which the answer should be provided. In the case of HeySQuAD, the prompt questions are the exact transcriptions of the audio input. Therefore, there is no need for the rephrasing stage ($R_{\text{text}} = R_{\text{0}}$). We also take 423K examples from Instructs2s~\cite{fang2024llama}: multi-turn responses are converted into the textual context, and only the last user query is provided in the spoken format.
The complete distribution across categories is provided in Table \ref{tab:dataset_overview}.

% cameo, index 553
\begin{table}[t]
    \centering
    \caption{Original and rephrased responses with and without limits from Qwen3-4B-Thinking-2507}
    \label{tab:rephrasing_example}
    \begin{tabularx}{\columnwidth}{>{\hsize=0.46\hsize}X|>{\hsize=1.5\hsize}X}
    \toprule
    Type &  Model response\\
    \midrule
    Original & We are \textbf{given} an audio snippet with \textbf{metadata}:
   [00:00-00:01] I'm on my way to the meeting. \textbf{(duration: 1s, emotion: Neutral, gender: Male, age: 34, language: English)}

 The prompt asks: "What is the tone of the speaker's voice in the audio?"

 \textbf{From the metadata provided}, we see: 
 emotion: Neutral...\\
    \midrule
    Rephrased  (no limit) & Okay, I \textbf{listen} to the audio clip. The speaker says 'I'm on my way to the meeting' during the first second (00:00-00:01). \textbf{I perceive the emotional state as neutral, the gender as male, the age as 34, and the language as English.}

The prompt asks: 'What is the tone of the speaker's voice in the audio?'...\\
    \midrule
    Rephrased (limit: $1536$) & I \textbf{hear} an audio clip that lasts 1 second, starting at 00:00 and ending at 00:01. The speaker says: "I'm on my way to the meeting." The prompt asks: "What is the tone of the speaker's voice in the audio?"

\textbf{From the audio I perceive}:
   emotional tone: neutral...\\
    \bottomrule
    \end{tabularx}
\end{table}

\subsection{ALARM}\label{sec:methodology_alarm}

Our audio reasoning model, ALARM, consists of a frozen RLM $Q_r$ augmented with trainable audio adapters and frozen encoders, as shown in Figure~\ref{fig:training_overview}. The adapters map encoder features into the input embedding space of $Q_r$, enabling the backbone to interpret audio representations and produce the desired outputs. Thus, adapter design and encoder choice are critical. Prior ALM systems~\cite{lu2025desta2, lu2025developing, xu2025qwen2, chu2024qwen2, yao2024minicpm, li2025megrez} largely rely on Whisper~\cite{radford2023robust} as a single encoder for both speech and non-speech audio. Since Whisper is optimized for ASR, its embeddings are not optimal for music and general audio understanding. Stacking multiple adapters on top of a single encoder, as in GAMA~\cite{ghosh2024gama}, cannot compensate for the encoder's suboptimal representations of non-target audio types, since the fundamental limitation imposed by the encoder's training data distribution persists. We therefore adopt a multi-encoder design: Whisper for speech, W2V-BERT-2.0~\cite{barrault2023seamless} for richer auditory cues from large-scale pretraining, MuQ~\cite{zhu2025muq} for music, and SSLAM~\cite{alex2025sslam} for general audio (sound), leveraging their strong domain-specific performance. However, having information from various encoders significantly increases the memory and computational cost if naively concatenated across the time dimension. Simple concatenation across feature-dimension, e.g. in SALMONN~\cite{tang2024salmonn}, is also suboptimal because it has limited ability to capture both cross-encoder interactions and within-encoder feature dependencies.

We propose three alternative approaches for improved feature fusion across multiple encoders. The first, \emph{ALARM-CA}, stacks consecutive cross-attention fusion modules, with each module refining the representation produced by the previous one conditioned on the input. The second, \emph{ALARM-P}, treats Whisper’s features as the primary stream and uses three separate Perceivers~\cite{jaegle2022perceiver} to compress the outputs of the other encoders into a fixed set of embeddings; these embeddings are then prepended as a prefix to the Whisper features. In the third approach, we create an ensemble encoding based on \emph{ALARM-CA} and Whisper, denoted as \emph{ALARM-E}. The first two approaches operate at a reduced token rate of 25 Hz (vs. the default 175 Hz), which lowers overall computation and memory costs. \emph{ALARM-E}, on the other hand, operates at 50 Hz and reveals an interesting trade-off between computational cost and performance.

In particular, given $(A_{\text{audio}}^{i}, P^{i}, R_{\text{text}}^{i}) \in \mathcal{D}$, each audio encoder $G_m(\cdot)$ encodes the input audio signal into a sequence of frame-level features
\[
f_{m}^{i} = G_m(A_{\text{audio}}^{i}) \in \mathbb{R}^{T_m \times D_m},
\]
where $T_m$ and $D_m$ denote the temporal resolution and feature dimensionality of encoder $G_m$, respectively.

Since different encoder layers capture complementary information, we aggregate multi-layer representations using a trainable weighted average. That is, for each encoder $G_m$ (we use $M=4$ encoders), we extract hidden states from a selected set of layers $\mathcal{L}_m$ and compute
\[
\bar{f}_{m}^{i} = \sum_{\ell \in \mathcal{L}_m} \alpha_{m,\ell}\, f_{m,\ell}^{i},
\qquad
\alpha_{m,\ell} = \frac{\exp(w_{m,\ell})}{\sum_{\ell' \in \mathcal{L}_m} \exp(w_{m,\ell'})},
\]
where $f_{m,\ell}^{i}$ denotes the hidden representation from layer $\ell$ of encoder $G_m$, and $\{w_{m,\ell}\}$ are learnable scalar parameters.

The resulting features $\bar{f}_{m}$ for each encoder is temporally compressed to a token rate of 25~Hz and mapped to the embedding space of the frozen RLM $Q_r$ using a modality adapter $\Theta_m(\cdot)$ as:
\[
x_{m}^{i} = \Theta_m(\bar{f}_{m}^{i}) \in \mathbb{R}^{\frac{T_m}{2} \times D_m}.
\]
For most encoders, the adapter $\Theta_m(\cdot)$ consists of a two-layer convolutional network that downsamples the feature sequence from 50~Hz to 25~Hz. For the MuQ encoder, however, $\Theta_m(\cdot)$ is implemented as a multilayer perceptron (MLP), since its default token rate is already 25~Hz. The adapted features $\{x_m^{i}\}_{m=1}^{M}$ are then integrated using the aforementioned approaches, as elaborated below.

\subsubsection{ALARM-CA}

In this method, fusion is performed using a stack of cross-attention blocks. At each stage, the output of the primary encoder serves as the query and is fused with the adapted features from another encoder via cross attention. We designate Whisper as the primary encoder due to its content-focused pretraining and denote its adapted features by $x_{\mathrm{wh}}^{i}$. The initial query is defined as
\[
h_{0}^{i} = x_{\mathrm{wh}}^{i}.
\]
The fusion is applied sequentially following a fixed order of encoders:
\[
\text{Whisper} \rightarrow \text{W2V-BERT-2.0} \rightarrow \text{MuQ} \rightarrow \text{SSLAM}.
\]
For each fusion stage $m \in \{1,2,3\}$, the output of the previous stage is used as the query and fused with the adapted features of the next encoder:
\[
h_{k}^{i} = \mathcal{A}_c\left(h_{m-1}^{i}, x_{m}^{i}\right),
\]
where $\mathcal{A}_c(\cdot,\cdot)$ denotes a $2$-layer cross-attention block.
The final fused representation $h_{3}^{i}$ is provided to the frozen reasoning language model $\mathcal{Q}_r$. Conditioned on $h_{3}^{i}$ and the instruction prompt $P^{i}$, $\mathcal{Q}_r$ is trained autoregressively to predict the target sequence $R_{\text{text}}^{i}$ using the cross-entropy loss:
\[
\mathcal{L}
= - \sum_{t} \log p_{\mathcal{Q}_r}\left(r_{t}^{i} \mid r_{<t}^{i}, h_{3}^{i}, P^{i}\right).
\]
We prepend the final audio features $h_{3}^{i}$ to the input sequence before the textual query $P^{i}$, as this ordering has been shown to improve generalization~\cite{fan2025alignformer}. In addition, two trainable continuous vectors are inserted before and after the audio features, respectively, to explicitly separate the audio and text modalities. Unlike many previous approaches \cite{lu2024desta, lu2025desta2, lu2025developing}, leveraging rich and continuous input representations avoids the dependency on voice activity detection (VAD) and mitigates the limitations related to purely text-based inputs in general audio understanding.

\subsubsection{ALARM-P}\label{sec:alarm_p}

In this method, the Whisper embeddings, $x_{\mathrm{wh}}^{i}$, obtained after the adapter with a 25Hz token rate, are used as the primary input to $\mathcal{Q}_r$, while all the complementary features from the other encoders are incorporated as a short fixed-length prefix of size $60$. We choose Whisper as the primary encoder because our experiments (see Section~\ref{sec:results}) indicate that, under the ALARM-CA approach, Whisper representations become less informative for speech and content-related tasks. Feeding them directly into $\mathcal{Q}_r$ therefore preserves their availability while still allowing them to be augmented with additional encoder features.

To this end, we first apply three separate Perceiver~\cite{jaegle2022perceiver} modules to the features $\bar{f}_{m}^{i} \in \mathbb{R}^{T_m \times D_m}$ extracted from W2V-BERT-2.0, SSLAM, and MuQ, transforming them into compressed fixed token sequences $p_{m}^{i} \in \mathbb{R}^{20 \times D_m}$ as follows:
\[
p_{m}^{i} = \mathcal{A}_{p}^{m}\left(\bar{f}_{m}^{i}\right),
\]
where $\mathcal{A}_{p}^{m}(\cdot)$ denotes the Perceiver module associated with encoder $G_m(\cdot)$. In this setup, the adapter $\Theta_m(\cdot)$ is no longer required, since $\mathcal{A}_{p}^{m}(\cdot)$ performs both temporal compression and feature adaptation.

The resulting tokens $p_{m}^{i}$ are concatenated with the Whisper features $x_{\mathrm{wh}}^{i}$ to form
\[
[p_{\text{W2V-BERT-2.0}}^{i},\, p_{\text{MuQ}}^{i},\, p_{\text{SSLAM}}^{i},\, x_{\mathrm{wh}}^{i}],
\]
which are then passed to $\mathcal{Q}_r$. Additionally, two trainable continuous embeddings are inserted before and after the Perceiver tokens to explicitly separate them from the Whisper sequence. The rest of the training pipeline is similar to ALARM-CA.

\subsubsection{ALARM-E}\label{sec:alarm_e}
Similar to \emph{ALARM-P}, the adapted Whisper encodings $x_{\mathrm{wh}}^{i}$ (25 Hz) are used as the primary features. However, instead of compressing the additional features from other encoders into a fixed number of tokens, we allow more expressive representations by concatenating $x_{\mathrm{wh}}^{i}$ with $h^i_3$ from \emph{ALARM-CA} along the time dimension, forming a 50~Hz frame rate. It should be noted that, a 50~Hz token rate compression is still superior over naively concatenating all encoders (175~Hz). To encourage the model to attend to distinct aspects of the audio signal, we introduce an auxiliary instruction $I_{\text{e}}$ that prompts the model to process the audio in two passes, each focusing on different characteristics from ALARM-CA and Whisper embeddings. The resulting audio prompt for \emph{ALARM-E} is as follows:

\[
[I_{\text{e}},\, I_1,\, h^i_3,\, I_2,\, x_{\mathrm{wh}}^{i}],
\]
where $I_1$ and $I_2$ are text prompts guiding the model for better feature separations. Since \emph{ALARM-E} is an \emph{inference-time-only} strategy, no training with 50Hz token rate is required. Each individual fusion module is trained at a 25Hz token rate, while the joint fusion module operates at 50Hz.

\section{Experimental Setup}\label{sec:exp_setup}

We use pretrained Qwen3-30B-A3B-Instruct-2507-FP8~\cite{yang2025qwen3} for the dataset collection and Qwen3-4B-Thinking-2507~\cite{yang2025qwen3} as our backbone RLM for the ALM.

The convolutional adapters $\Theta_m(\cdot)$ consist of two convolutional layers with kernel sizes $3$ and $4$, separated by a GELU activation and a normalization layer. The MLP adapter used for MuQ comprises two linear layers with a hidden dimension of $1024$ and a GELU activation in between. The outputs of all adapters and Perceiver modules are projected to the RLM input dimension via simple linear layers whenever required. When performing layer-wise feature aggregation (see Section~\ref{sec:methodology_alarm}), we select layers $\{8,16,24,32\}$ for Whisper, $\{7,11,16,21,25\}$ for W2V-BERT~2.0, $\{5,9,13\}$ for MuQ, and $\{4,8,12\}$ for SSLAM. This selection is motivated by previous work \cite{pasad2023comparative} that highlights that audio foundation models encode different types of information in their lower, middle, and higher layers. For the multi-encoder setup, we found it beneficial to initialize ALARM-CA adapters with the weights from our single-encoder systems: AL-Whisper-R, AL-W2VBERT2-R, AL-MuQ-R, and AL-SSLAM-R (see Section \ref{sec:results}). We freeze everything except for the fusion module and train the system for 1 epoch on all data. In contrast, ALARM-P takes the frozen pretrained AL-Whisper-Instruct-R while training the perceivers from scratch. AL-Whisper-Instruct-R is also used for Whisper embeddings in ALARM-E.

We split the compiled ALARM-corpora into training and validation sets. For each subset, we construct the validation partition by randomly sampling $10\%$ of its instances, except for the speech-instruction subset, for which we use the official HeySQuAD validation split.

All of our models are trained with an effective batch size of 64 for 2 epochs (unless specified otherwise) using AdamW optimizer with 1500 warm-up steps and a maximum learning rate of $10^{-4}$ with a cosine annealing scheduling on 4 H200 NVIDIA GPUs.

\section{Results}\label{sec:results}

\begin{table*}[!htp]
    \centering
    \caption{Accuracies on the MMSU benchmark. We further report the text LLM backbone size, whether it is kept frozen ($\checkmark$) or modified ($\times$), and the training data volume. Best results are in bold, second best underlined, third best are in italic.}
    \label{tab:mmsu}
    \resizebox{\linewidth}{!}{
    \begin{tabular}{l|ccc|ccc}
    \toprule
    \multirow{2}{*}{Model} & \multirow{2}{*}{Size} & \multirow{2}{*}{\shortstack{No LLM \\ Modification}} & \multirow{2}{*}{\shortstack{$\#$ Samples $|$ $\#$ Hours $|$ $\#$ Tokens}} & \multicolumn{3}{c|}{MMSU} \\
    \cmidrule{5-7}
    & & & & Perception $\uparrow$ & Reasoning $\uparrow$ & Overall $\uparrow$ \\
    \midrule
    GPT-4o Audio~\cite{hurst2024gpt} & $-$ & $-$ & $-$ & $39.7$ & $71.2$ & $56.4$ \\
    Gemini-1.5-Pro~\cite{team2024gemini}  & $-$ & $-$ & $-$ & $\textit{46.3}$ & $76.1$ & $60.7$ \\
    \midrule
    LTU~\cite{gong2024listen} & 7B & $\times$ &  $5.6$M $|$ $-$ $|$ $-$  & $20.8$ & $24.4$ & $22.6$ \\
    SALMONN~\cite{tang2024salmonn} & 7B & $\times$ & $2.3$M $|$ $4.4$K $|-$  & $29.8$ & $30.0$ & $30.1$ \\
    Baichuan-Omni-1.5~\cite{li2025baichuan} & 7B & $\times$ & $-|$ $887$K $|-$ & $35.4$ & $67.2$ & $50.6$ \\
    Qwen2-Audio-Instruct~\cite{chu2024qwen2} & 7B & $\times$ & $-$ $|$ $320$K $|$ $-$  & $39.0$ & $68.9$ & $53.3$ \\
    Step-Audio 2 mini~\cite{wu2025step} & 7B & $\times$ &  $-|-|$ $0.2T$ & $42.7$ & $72.6$ & $57.2$ \\
    MiniCPM-o-2.6~\cite{yao2024minicpm} & 7B & $\times$ & $-$ & $40.5$ & $73.6$ & $56.5$ \\
    MiMo-Audio-7B-Instruct + Think~\cite{zhang2025mimo} & 7B & $\times$ & $-|-|$ $2.4$T & $\bm{51.7}$ & $74.8$ & $\bm{62.9}$ \\
    Kimi-Audio~\cite{ding2025kimi} & 7B & $\times$ & $-$ $|$ $13.3$M $|$ $585$B & $43.5$ & $76.0$ & $59.3$ \\
    MiMo-Audio-7B-Instruct~\cite{zhang2025mimo} & 7B & $\times$ & $-|-|$ $2.4$T & $\underline{46.9}$ & $\textit{76.7}$ & $\underline{61.7}$ \\
    Qwen2.5-Omni~\cite{xu2025qwen2} & 7B & $\times$ & $-$ $|$ $-$ $|$ $0.3$T & $42.5$ & $\bm{79.8}$ & $60.6$ \\
    \midrule
    Phi-4-Multimodal~\cite{abouelenin2025phi} & 4B & $\times$ & $>28$M $|$ $>2$M $|$ $-$ & $33.4$ & $57.6$ & $45.0$ \\
    Megrez-3B-Omni~\cite{li2025megrez} & 2B &  $\times$ & $-$ & $32.5$ & $67.1$ & $49.0$ \\
    Qwen2.5-Omni~\cite{xu2025qwen2} & 3B & $\times$ &  $-$ $|$ $-$ $|$ $0.3$T & $42.4$ & $72.8$ & $56.8$ \\
    \midrule
    AL-Whisper-Instruct-R  & 4B & \checkmark & $0.5$M $|$ $1$K $|$ $91$M & $38.2$ & $73.6$ & $55.3$\\
    \midrule
    ALARM-CA & 4B & \checkmark & $5.5$M $|$ $17$K $|$ $1.5$B & $39.6$ & $68.3$ & $53.5$\\
    ALARM-P & 4B & \checkmark & $5.5$M $|$ $17$K $|$ $1.5$B & $38.4$ & $74.2$ & $55.8$\\
    \midrule
    ALARM-E & 4B & \checkmark & $5.5$M $|$ $17$K $|$ $1.5$B & $45.4$ & $\underline{78.3}$ & $\textit{61.3}$\\
    \bottomrule
    \end{tabular}
    }
\end{table*}

We evaluate our models across a diverse set of speech and general audio understanding benchmarks. We first focus on speech understanding tasks and examine the model’s ability to preserve a balance between its original pre-trained knowledge and the knowledge acquired through our modality alignment approach. We then assess general audio understanding performance, covering music, speech, and environmental sound domains. Finally, we conduct an ablation study to analyze the contribution of each individual encoder and evaluate the model on a more fine-grained benchmark.

\subsection{Speech Understanding and Textual Generalization}\label{sec:results_mmsu}

For spoken language understanding, we evaluate on MMSU~\cite{wang2025mmsu}, which includes 47 perception and reasoning tasks across 5000 samples covering both linguistic and paralinguistic information. The reasoning subset is particularly relevant to our objectives. Table~\ref{tab:mmsu} reports the accuracy of our models and recent ALMs.

On the reasoning partition, ALARM-P outperforms GPT-4o-Audio~\cite{hurst2024gpt} and most competing ALMs, including all models of comparable or smaller size and the majority of larger ones, by a substantial margin. In contrast, ALARM-CA exhibits a notable drop in reasoning accuracy. We attribute this behavior to the loss of ASR features from Whisper during the compression process. To validate this hypothesis, we train a Whisper-only model, AL-Whisper-Instruct-R, on the speech instruction subset of the ALARM corpora. As shown in Table~\ref{tab:mmsu}, it significantly outperforms ALARM-CA on the reasoning partition, highlighting the importance of strong content extraction.

By design, AL-Whisper-Instruct-R is less effective on perceptual tasks. ALARM-P/CA improves this score by incorporating additional features; although Table~\ref{tab:mmsu} confirms their benefit, the improvement in speech understanding remains limited. ALARM-E, on the other hand provides better input signals by concatenating ALARM-CA and AL-Whisper-Instruct-R embeddings. This complementary combination substantially enhances both perceptual and reasoning performance. This is a strong indication that the ASR features based on Whisper, benefit from complementary features from other audio encoders. 

ALARM-E achieves the second-best result on the MMSU reasoning partition, trailing the 7B Qwen2.5-Omni~\cite{xu2025qwen2} by only $1.5\%$. It also surpasses Qwen2.5-Omni by $2.9\%$ on perception tasks and by $0.7\%$ overall. Its total MMSU score ranks third, behind only the MiMo~\cite{zhang2025mimo}, which benefits from much larger training data (2.4T tokens). Overall, ALARM-E demonstrates superior reasoning performance, outperforming even Gemini-1.5-Pro~\cite{team2024gemini} and MiMo-Audio-7B-Instruct-Think, the only other ALM with explicit chain-of-thought reasoning.

Unlike MiMo and Qwen2.5-Omni, which fully fine-tune their LLM backbones on large-scale datasets at substantially higher cost, ALARM-E keeps the RLM \emph{frozen} and uses over $200\times$ fewer audio tokens. Moreover, freezing the LLM preserves its original text capabilities. As shown in Table~\ref{tab:text_benchmarks}, multimodal fine-tuning drastically degrades the performance on purely textual tasks. In contrast, ALARM models do not modify the backbone weights and retain full text performance, striking a balance between the two modalities.

\subsection{General Audio Understanding}\label{sec:results_mmau_mmar}

\begin{table*}[!h]
    \centering
    \caption{Comparison of text LLMs/RLMs and their corresponding multimodal variants on different text benchmarks.}
    \label{tab:text_benchmarks}
    \resizebox{\linewidth}{!}{
    \begin{tabular}{l|c|ccccc}
    \toprule
    Model & Size & MMLU-Pro~\cite{wang2024mmlu} $\uparrow$ & MMLU-Redux~\cite{gema2025we} $\uparrow$ & GPQA~\cite{rein2024gpqa} $\uparrow$ & GSM8K~\cite{cobbe2021training} $\uparrow$ & HumanEval~\cite{chen2021evaluating} $\uparrow$\\
    \midrule
    Megrez-3B-Instruct~\cite{li2025megrez} & 2B & $46.1$ & $-$ & $-$ & $65.5$ & $78.7$ \\
    Megrez-3B-Omni~\cite{li2025megrez} & 2B & $45.2$ & $-$ & $-$ & $63.8$ & $72.6$ \\
    \midrule
    Qwen2.5~\cite{yang2024qwen2_5} & 7B & $56.3$ & $75.4$ & $36.4$ & $91.6$ & $84.8$\\
    Qwen2.5-Omni~\cite{xu2025qwen2} & 7B & $47.0$ & $71.0$ & $30.8$ & $88.7$ & $78.7$ \\
    \midrule
    Qwen3-4B-Thinking-2507~\cite{yang2025qwen3} & 4B & $74.0$ & $86.1$ & $65.8$ & $-$ & $-$ \\
    ALARM (ours, any variant) & 4B & $74.0$ & $86.1$ & $65.8$ & $-$ & $-$\\
    \bottomrule
    \end{tabular}
    }
\end{table*}

\begin{table*}[!ht]
    \centering
    \caption{Results on MMAU and MMAR benchmarks. We further report the text LLM backbone size, whether it is kept frozen ($\checkmark$) or modified ($\times$), and the training data volume. Best results are in bold, second best underlined.}
    \label{tab:mmau_mmar}
    \resizebox{\linewidth}{!}{
    \begin{tabular}{l|ccc|cccc|c}
    \toprule
    \multirow{2}{*}{Model} & \multirow{2}{*}{Size} & \multirow{2}{*}{\shortstack{No LLM \\ Modification}} & \multirow{2}{*}{\shortstack{$\#$ Samples $|$ \\ $\#$ Hours $|$ $\#$ Tokens}} & \multicolumn{4}{c|}{MMAU-v05.15.25 (Test-mini / Test)} & MMAR \\
    \cmidrule{5-9}
    & & & & Sound $\uparrow$ & Music $\uparrow$ & Speech $\uparrow$ & Mean $\uparrow$ & Mean $\uparrow$\\
    \midrule
    GPT-4o mini Audio~\cite{hurst2024gpt} & $-$ & $-$ & $-$ & $50.8$ / $49.7$ & $39.2$ / $36.0$ & $69.1$ / $67.5$ & $53.0$ / $51.0$ & $50.6$ \\
    GPT-4o Audio~\cite{hurst2024gpt} & $-$ & $-$ & $-$ & $64.6$ / $63.2$ & $56.3$ / $49.9$ & $66.7$ / $69.3$ & $62.5$ / $60.8$ & $63.5$ \\
    \midrule
    LTU~\cite{gong2024listen} & 7B & $\times$ & $5.6$M $|$ $-$ $|$ $-$ & $20.4$ / $20.7$ & $16.0$ / $15.7$ & $15.9$ / $15.3$ & $17.4$ / $17.2$ & $19.2$ \\
    GAMA~\cite{ghosh2024gama} & 7B & $\times$ & $6.3$M $|-|-$ & $31.8$ / $30.7$ & $17.7$ / $17.3$ & $12.9$ / $16.7$ & $20.8$ / $21.7$ & $26.5$ \\
    GAMA-IT~\cite{ghosh2024gama} & 7B & $\times$ & $6.5$M $|-|-$ & $30.9$ / $32.7$ & $26.7$ / $22.4$ & $10.8$ / $11.6$ & $22.8$ / $22.2$ & $17.4$ \\
    MU-LLaMA~\cite{liu2024music} & 7B & \checkmark & $0.1$M $|$ $76$ $|-$ & $33.0$ / $31.0$ & $32.3$ / $29.7$ & $17.4$ / $17.1$ & $27.6$ / $25.9$ & $13.9$ \\
    SALMONN~\cite{tang2024salmonn} & 13B & $\times$ & $2.3$M $|$ $4.4$K $|-$ & $41.1$ / $42.1$ & $37.1$ / $37.8$ & $26.4$ / $28.8$ & $34.9$ / $36.2$ & $33.2$ \\
    MusiLingo~\cite{deng2024musilingo} & 7B & \checkmark & $0.7$M $|-|-$ & $43.2$ / $41.9$ & $40.1$ / $41.2$ & $31.2$ / $31.7$ & $38.2$ / $38.3$ & $6.6$ \\
    Gemma 3n~\cite{gemma3n2025} & 6B2E & $-$ & $-$ & $51.4$ / $47.5$ & $52.1$ / $51.6$ & $52.2$ / $57.0$ & $51.7$ / $52.0$ & $-$ \\
    Gemma 3n~\cite{gemma3n2025} & 8B4E & $-$ & $-$ & $55.9$ / $50.2$ & $56.9$ / $53.2$ & $61.3$ / $62.1$ & $58.0$ / $55.2$ & $-$ \\
    Qwen2-Audio-Instruct~\cite{chu2024qwen2} & 7B & $\times$ & $-$ $|$ $320$K $|$ $-$ & $67.3$ / $61.2$ & $56.3$ / $55.7$ & $55.3$ / $55.4$ & $59.6$ / $57.4$ & $30.0$ \\
    \midrule
    DeSTA-2.5-Audio~\cite{lu2025desta2} & 8B & \checkmark & $5$M $|$ $7$K $|$ $-$ & $70.3$ / $66.9$ & $56.3$ / $57.1$ & $71.5$ / $71.9$ & $66.0$ / $65.2$ & $50.8$ \\
    Audio Reasoner~\cite{zhifei2025audio} & 8B & $\times$ & $>1.2$M $| > 320$K $|-$ & $67.9$ / $67.3$ & $69.2$ / $61.5$ & $66.1$ / $62.5$ & $67.7$ / $63.8$ & $36.8$ \\
    Qwen2.5-Omni~\cite{xu2025qwen2} & 7B & $\times$ & $-$ $|$ $-$ $|$ $0.3$T & $78.1$ / $76.8$ & $65.9$ / $67.3$ & $70.6$ / $68.9$ & $71.5$ / $71.0$ & $56.7$ \\
    Audio Flamingo 3~\cite{ghosh2025audio3} & 8B & $\times$ & $26.7$M $|$ $54.4$K $|$ $-$ & $79.6$ / $75.8$ & $74.0$ / $74.5$ & $66.4$ / $67.0$ & $73.3$ / $72.4$ & $58.5$ \\
    \midrule
    Phi-4-Multimodal~\cite{abouelenin2025phi} & 4B & $\times$ & $>28$M $|$ $>2$M $|$ $-$ & $65.5$ / $62.7$ & $64.37$ / $62.0$ & $67.27$ / $63.8$ & $66.70$ / $62.8$ & $-$ \\
    Audio Flamingo 2~\cite{ghosh2025audio2} & 3B & $\times$ &  $5.9$M $|$ $-$ $|$ $-$ & $71.5$ / $68.1$ & $71.0$ / $70.2$ & $44.7$ / $44.9$ & $62.4$ / $61.1$ & $21.9$ \\
    \midrule % single system
    AL-Whisper-Instruct-R & 4B & \checkmark & $0.5$M $|$ $1$K $|$ $91$M & $45.9$ / $49.9$ & $46.7$ / $49.7$ & $\underline{73.0}$ / $\underline{73.0}$ & $55.2$ / $57.5$ & $49.1$ \\
    AL-W2VBERT2-R & 4B & \checkmark & $0.5$M $|$ $0.6$K $|$ $58$M & $46.2$ / $46.9$ & $47.9$ / $46.0$ & $47.7$ / $48.2$ & $47.3$ / $47.0$ & $37.4$ \\
    AL-MuQ-R & 4B & \checkmark & $0.4$M $|$ $1.5$K $|$ $137$M & $50.5$ / $49.1$ & $53.9$ / $53.3$ & $40.84$ / $43.1$ & $48.4$ / $48.5$ & $36.0$ \\
    AL-SSLAM-R & 4B & \checkmark & $0.4$M $|$ $1.2$K $|$ $108$M & $61.3$ / $60.4$ & $46.7$ / $50.9$ & $45.9$ / $47.8$ & $51.3$ / $53.0$ & $33.8$ \\
    \midrule
    AL-Whisper-R & 4B & \checkmark & $5.5$M $|$ $17$K $|$ $1.5$B & $52.3$ / $55.6$ & $46.7$ / $49.7$ & $57.1$ / $58.8$ & $52.0$ / $54.7$ & $37.3$ \\
    \midrule
    ALARM-CA & 4B & \checkmark & $5.5$M $|$ $17$K $|$ $1.5$B & $66.4$ / $61.1$ & $57.2$ / $53.3$ & $60.1$ / $61.1$ & $61.2$ / $58.5$ & $40.6$ \\
    ALARM-P & 4B & \checkmark & $5.5$M $|$ $17$K $|$ $1.5$B & $58.6$ / $58.5$ & $52.7$ / $53.3$ & $68.8$ / $66.7$ & $60.0$ / $59.5$ & $43.5$ \\
    \midrule
    ALARM-E & 4B & \checkmark & $5.5$M $|$ $17$K $|$ $1.5$B & $64.0$ / $59.1$ & $54.8$ / $54.2$ & $\bm{77.2}$ / $\bm{73.7}$ & $65.3$ / $62.4$ & $48.7$ \\
    \bottomrule
    \end{tabular}
    }
\end{table*}

In this section, we evaluate ALARM on advanced audio understanding benchmarks. We use two datasets, namely MMAU~\cite{sakshi2025mmau} and MMAR~\cite{ma2025mmar}. MMAU (v05.15.25) comprises 27 expert-level, multi-step reasoning tasks across speech, sound, and music, with 1000 test-mini and 9000 test samples. MMAR~\cite{ma2025mmar} targets graduate-level, domain-specific reasoning and includes speech, sound, music, and mixed audio types across 1000 samples.

Table~\ref{tab:mmau_mmar} compares our models with recent ALMs. The MMAU results dovetail well with the findings on MMSU. In particular, the gap between AL-Whisper-Instruct-R and ALARM-CA on MMAU-Speech shows that cross-attention fusion in ALARM-CA underperforms on speech category while improving on sound and music categories.
In contrast, ALARM-P preserves speech understanding more effectively by separating perceiver tokens from Whisper tokens. However, projecting perceptual information onto a fixed token set still limits its performance on sound and music categories. ALARM-E combines the perceptual strengths of ALARM-CA with the content understanding of AL-Whisper-Instruct-R, leading to substantial gains on MMAU-Speech while maintaining strong sound and music performance.

Compared to other ALMs, ALARM-E achieves the highest open-source accuracy on MMAU-Speech and ranks top-3 among all the models, including both proprietary and open-access models, according to its current official leaderboard\footnote{\url{https://sakshi113.github.io/mmau\_homepage/\#leaderboard-v15-parsed}}. Notably, it surpasses DeSTA-2.5-Audio~\cite{lu2025desta2}, the previous open-source leader, by $5.7\%$ on the test-mini and $1.8\%$ on the full test speech subset.

In terms of overall MMAU score, ALARM-E outperforms most larger ALMs, including GPT-4o Audio, and Audio Flamingo 2~\cite{ghosh2025audio2}. It performs on par with the 4B Phi-4-Multimodal~\cite{abouelenin2025phi}, trading weaker music understanding for stronger speech processing. On MMAR, ALARM-E also surpasses most competing systems, including Audio Reasoner~\cite{zhifei2025audio}, despite its explicit chain-of-thought reasoning.

\begin{figure}
    \centering
    \includegraphics[width=\linewidth]{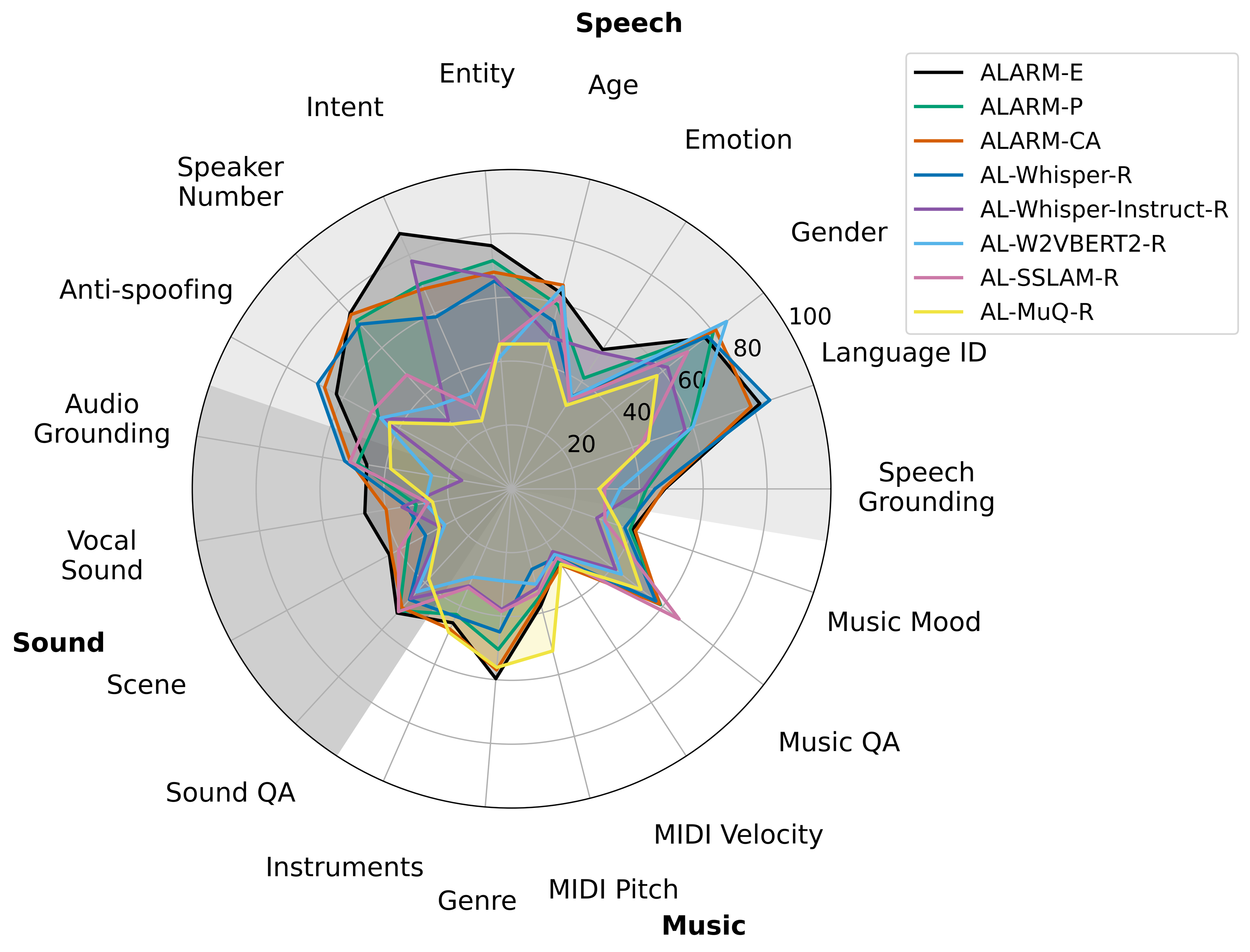}
    \caption{Comparison of our models on AIR-Bench benchmark across different tasks.}
    \label{fig:airbench}
\end{figure}

\subsection{Encoder Comparison and Fine-Grained Evaluation}\label{sec:results_multi_vs_single}

To assess the impact of the multi-encoder design, we implement single-encoder variants: AL-Whisper-R, AL-W2VBERT2-R, AL-MuQ-R, and AL-SSLAM-R. Each model employs only one encoder $m$, whose adapter features $x_m^i$ are directly fed into the RLM. AL-Whisper-R is trained on the \emph{full dataset}, while AL-Whisper-Instruct-R uses only the speech instruction subset. AL-W2VBERT2-R is trained on speech characteristics data (Cameo, Globe V3, VCTK, NISQA, Noisy VCTK), AL-MuQ-R on music data (Nsynth, SonicMaster), and AL-SSLAM-R on sound understanding datasets (Clotho, AudioCaps, ESC50, FSD50K, and $20\%$ of AudioSet). All models are trained for 2 epochs with an effective batch size of 32.

First, we compare the single-encoder systems with ALARM models on MMAU. Table~\ref{tab:mmau_mmar} shows that each single-encoder variant achieves strong performance in the MMAU categories aligned with its defined objective. For instance, AL-SSLAM-R, serving as a general audio backbone, exhibits high performance on MMAU-Sound, whereas AL-MuQ-R achieves stronger results in the music domain. Another interesting observation is that simply incorporating more diverse audio data is insufficient, as reflected by the performance of AL-Whisper-R (trained on the full dataset). Although its MMAU-Sound quality improves compared to the instruction-only checkpoint, it still underperforms relative to type-specific single-encoder models. Moreover, modest gains in sound understanding come at the cost of substantially degraded speech understanding. In contrast, employing multiple encoders (ALARM-CA) enables performance that surpass those from individual encoders in most of the categories, and further ensembling in ALARM-E yields additional improvements. These results give support to the effectiveness of our fusion strategies that are able to boost the overall performance while keeping the token rate lower than all the encoders combined.

We further analyze the behavior of our single- and multi-encoder systems on downstream tasks using AIR-Bench~\cite{yang2024air}, which spans 19 tasks across sound, music, and speech. Figure~\ref{fig:airbench} presents a radar plot comparing all proposed models.

Consistent with previous findings, the content-focused AL-Whisper-Instruct-R demonstrates strong zero-shot generalization, performing well on several speech characteristic extraction tasks. However, it is clearly outperformed by AL-MuQ-R and AL-SSLAM-R on music and sound categories. Notably, incorporating the W2VBERT-2.0 encoder enhances speech understanding, as AL-W2VBERT-2.0-R surpasses AL-Whisper-Instruct-R and achieves performance comparable to multi-encoder ALARM models on age and gender detection. 

Training AL-Whisper-R on the full dataset instead of only speech instructions substantially improves performance across most AIR-Bench tasks, except for Intent Classification and Emotion Detection. This aligns well with its improved MMAU-Sound results over AL-Whisper-Instruct-R in Table~\ref{tab:mmau_mmar}.

Finally, incorporating multiple encoders further improves performance over all single-encoder models across all tasks, except for MIDI pitch detection, where AL-MuQ-R remains superior. This underscores the importance of multi-encoder designs for comprehensive audio understanding beyond speech. Differences between ALARM-P and ALARM-CA are also reflected in their task performance: ALARM-P excels in speech-related tasks, while ALARM-CA performs better on audio and music tasks, consistent with the MMAU and MMAR results in Table~\ref{tab:mmau_mmar}. The same trend holds for ALARM-E, which delivers the strongest overall results among the ALARM variants while maintaining an effective balance between performance and information compression.

\section{Conclusion}\label{sec:conclusion}

In this paper, we presented ALARM, a framework for integrating audio understanding into large language models with chain-of-thought reasoning capabilities. Our self-rephrasing mechanism extends the self-generation paradigm to reasoning LLMs, producing training targets with natural thinking traces while preserving distributional alignment with the frozen backbone. By removing the dependency on ASR-derived inputs and fusing multiple audio encoders through cross-attention and Perceiver-based techniques, our approach enables robust understanding of both vocal and non-vocal audio signals within compact feature representations. Our feature-ensembling method (ALARM-E) allows our 4B-parameter ALM to achieve top-tier results on reasoning-oriented benchmarks, such as MMSU and speech-MMAU, outperforming models of comparable and even larger scale, while retaining the backbone's textual capabilities at substantially lower training cost. Notably, our observations confirm that while fine-tuning the text-based LLM improves the performance on audio-related tasks, it leads to degradation of its pre-trained textual knowledge. In contrast, our ALARM-E model demonstrates that medium-scale training on paired text–audio data is sufficient for strong speech understanding, even without fine-tuning the LLM. Furthermore, our analysis of encoder compression and fusion reveals that while augmenting Whisper embeddings with complementary audio features is beneficial, the fusion mechanism must be meticulously designed to maintain robustness across diverse audio domains.

\bibliographystyle{IEEEtran}
\bibliography{mybib}

\end{document}